\documentclass{article}
\usepackage{neurips_2023}

\usepackage[utf8]{inputenc} 
\usepackage[T1]{fontenc}    
\usepackage{hyperref}       
\usepackage{url}            
\usepackage{booktabs}       
\usepackage{amsfonts}       
\usepackage{nicefrac}       
\usepackage{microtype}      
\usepackage{xcolor}         
\usepackage[pdftex]{graphicx}
\usepackage{wrapfig}

\title{Forecasting Post-Wildfire Vegetation Recovery in California using a Convolutional Long Short-Term Memory Tensor Regression Network}

\author{%
  Jiahe Liu \\
  Western Connecticut State University\\
  Danbury, CT 06810 \\
  \texttt{jackliupenny@gmail.com} \\
   \And
   Xiaodi Wang \\
   Western Connecticut State University \\
   Danbury, CT 06810 \\
   \texttt{xiaodiwang1@yahoo.com} \\
}

\begin{document}

\maketitle

\begin{abstract}

The study of post-wildfire plant regrowth is essential for developing successful ecosystem recovery strategies. Prior research mainly examines key ecological and biogeographical factors influencing post-fire succession. This research proposes a novel approach for predicting and analyzing post-fire plant recovery. We develop a Convolutional Long Short-Term Memory Tensor Regression (ConvLSTMTR) network that predicts future Normalized Difference Vegetation Index (NDVI) based on short-term plant growth data after fire containment. The model is trained and tested on 104 major California wildfires occurring between 2013 and 2020, each with burn areas exceeding 3000 acres. The integration of ConvLSTM with tensor regression enables the calculation of an overall logistic growth rate $k$ using predicted NDVI. Overall, our $k$-value predictions demonstrate impressive performance, with 50\% of predictions exhibiting an absolute error of 0.12 or less, and 75\% having an error of 0.24 or less. Finally, we employ Uniform Manifold Approximation and Projection (UMAP) and KNN clustering to identify recovery trends, offering insights into regions with varying rates of recovery. This study pioneers the combined use of tensor regression and ConvLSTM, and introduces the application of UMAP for clustering similar wildfires. This advances predictive ecological modeling and could inform future post-fire vegetation management strategies.

\end{abstract}

\section{Introduction}

Wildfires are one of the most damaging and costly natural disasters in the United States. Data released by the National Interagency Fire Center (NIFC) indicates that there has been a gradual rise in the moving averages of both total burned acreage and burned acreage per fire over the past few decades [1]. From 2012 to 2022, an average of 7.3 million acres of vegetation were burned by wildfires annually in the U.S., with 3 extreme years from 2015 to 2021 where the yearly burned acreage exceeded 10 million acres, surpassing all pre-2015 records in modern data collection [1]. This results in heightened costs, ranging from direct costs such as firefighting efforts and property damage, to indirect costs such as the resulting implications for water supplies and public health. In the U.S., dedicated federal funding for fire suppression alone reached \$3.65 billion as of 2020 [2].

Across the US, California is by far impacted the most by wildfires, which has had more than 59,000 structures destroyed by wildfires between 2005 and 2020 [2]. Out of the 20 costliest and most destructive fires in California's history, 10 of them have taken place since 2015 [3]. Because California is a hotspot for major wildfires, it serves as an area of interest for our study.
 
Post-fire succession refers to the natural sequence of plant and ecosystem changes that occur after a wildfire, including the colonization of plant species, regrowth of vegetation, and reestablishment of a {
\parfillskip=0pt
\parskip=0pt
\par}
\begin{wrapfigure}[19]{R}{0.45\textwidth}
  \centering
  \includegraphics[width=1.\linewidth]{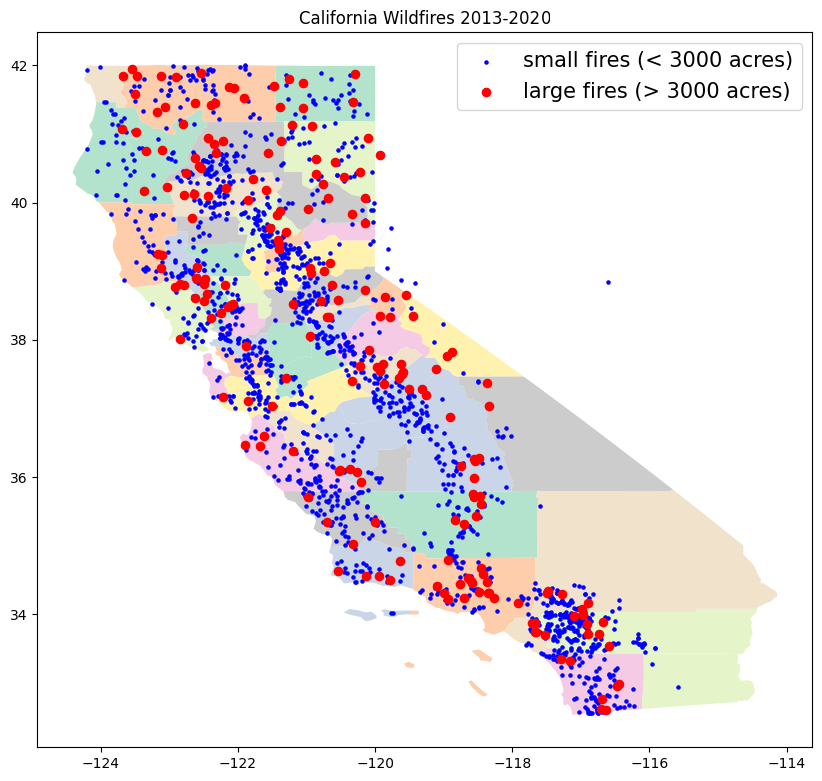}
  \caption{A scatter plot of the coordinates of every wildfire collected in this study. The large wildfires represented by red dots are included in our dataset, while the small wildfires represented by blue dots are not.}
  \label{fig:distribution}
\end{wrapfigure}
stable ecosystem. Human intervention may be required to hasten the post-fire succession of severely damaged forests, as it can be difficult for these areas to naturally recover [3]. Therefore, the study of post-fire plant regeneration plays a crucial role in guiding decision-making in conservation efforts.

In this study, a Convolutional Long Short-Term Memory (ConvLSTM) network was utilized to forecast future vegetation levels up to two years after the containment of large wildfires in California. Then, we utilized logistic curve fitting and tensor regression to derive logistic curve parameters, namely a growth rate $k$ for each fire, creating a combined ConvLSTM Tensor Regression (ConvLSTMTR) network. Comparing the fitted $k$-values on actual post-wildfire vegetation levels with the fitted $k$-values on our ConvLSTMTR results, we found that the framework performed consistently well on out-of-sample wildfires, with 50\% of predictions having an absolute error of 0.12 or less. 

\subsection{Application Context}

Many studies around post-wildfire forest regeneration have been conducted by analyzing specific species of trees or vegetation, allowing us to understand it from an ecological and biogeographical perspective. The severity of a wildfire can significantly damage seed availability [4, 5] and forest soils, slowing the rate of recovery [6]. These factors all impact the duration of post-wildfire regeneration, which can take years to fully occur [8, 9]. Deep learning has been explored in a study that forecasts fire danger maps [10]. However, knowing how to respond to the aftermath of a wildfire is just as important, especially as they have been increasing in severity and burn area. We propose a framework for long-term post-wildfire recovery analysis, which will provide the projections needed for disaster response communities to know which regions of vegetation need the most resources for recovery.

\section{Data and Methods}

\subsection{Study Area}
We selected the state of California as our study area because of its frequent and major wildfires, generating a comprehensive dataset that enhances the depth of our analysis. Data points including coordinates, burn areas, and fire start and end dates for wildfires from 2013 were publicly available and retrieved from the California Department of Forestry and Fire Protection (CAL FIRE). Our goal is to analyze large wildfires whose burn areas were significant enough so that clear vegetation growth patterns could be analyzed, so we excluded wildfires with a total burn area of less than $3,000$ acres (12.14 km$^2$), leaving our dataset with 104 wildfires. Figure \ref{fig:distribution} shows the distribution of all wildfires collected and highlights the large wildfires selected in this study.

\subsection{Data Collection}
In this study, global spatial indices were collected from NASA's Moderate Resolution Imaging Spectroradiometer (MODIS) to represent key features of post-wildfire recovery: vegetation levels, surface temperature, precipitation, and a mask of vegetation area affected by each wildfire. The target variable to be predicted was the Normalized Difference Vegetation Index (NDVI) collected from MOD13Q1 [11]. NDVI is a commonly used index to quantify vegetation greenness and biomass density that is derived from Landsat surface reflectance indices. As a supplement to NDVI, the Enhanced Vegetation Index (EVI) was also collected from MOD13Q1, as it exhibits greater sensitivity to areas with high biomass than NDVI. This gave our model a wider and more accurate range of vegetation values. Both NDVI and EVI have a range of $-0.2$ to $1$. 

Since temperature plays a major role in plant growth [12, 7], the Land Surface Temperature (LST) index was collected from MOD11A2 [13]. A fire mask was also collected from MOD14A2 to indicate the severity of wildfires in affected areas. Lastly, daily precipitation data was acquired from NASA's Integrated Multi-satellite Retrievals for GPM (IMERG) [14]. The length of each time series was set to 25 time steps with an interval of 32-day interval, starting from the month when each wildfire was marked as contained. For each of these indices, a 50 by 50-pixel bounding box was extracted with the coordinates of each wildfire's origin as the center.

\subsection{Data Preprocessing}

\paragraph{NDVI Value Imputation}

As MOD13Q1 derives its NDVI calculations from satellite-detected reflectance signals, the quality of its data is significantly impacted by fluctuations in snow and ice cover, cloud shadows, and different viewing angles. This would introduce noise in our dataset, disrupting our model's ability to accurately capture temporal trends and spatial patterns in post-fire vegetation recovery. To address this issue, we extracted a pixel reliability channel from MOD13Q1 to assign a quality rank to each pixel. Pixel reliability values of 2 or 3 indicate that these regions were affected by snow, ice, or cloud cover, so we assigned NaN values to replace unreliable pixels. We then utilized KNN imputation to fill in missing values, allowing us to preserve variability by taking neighbors from both spatial and temporal dimensions.

\paragraph{NDVI De-Seasonalization and Normalization}

NDVI values exhibit noticeable seasonality trends due to the natural cycles and processes that affect plant growth. These trends can differ in both magnitude and time frame, resulting in the need for normalization across all NDVI frames collected. We collected a series of monthly reference NDVI frames up to one year before the start of each wildfire. Then, for every post-wildfire NDVI frame, its pixel values would be divided by the corresponding values from the reference NDVI frame of the same month. This approach transformed our model's task from forecasting absolute and fluctuating NDVI values to an NDVI ratio, where a value of 1 signifies pre-wildfire vegetation levels.

\subsection{Convolutional LSTM Model Architecture}
\begin{figure}[t] 
  \centering
  \includegraphics[width=1.\linewidth]{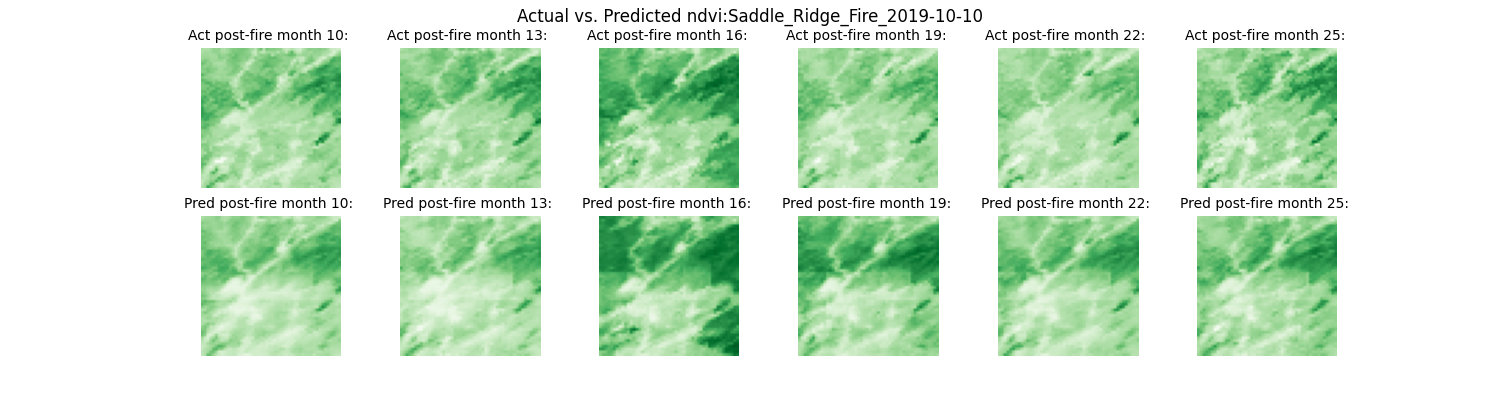}
  \caption{A comparison between the actual post-fire NDVI frames (top) and our model’s forecasted NDVI frames (bottom) for the Saddle Ridge Fire of 2019.}
  \label{fig:saddle}
\end{figure}

The model of choice for this study was a Convolutional Long Short-Term Memory Network (ConvLSTM). The ConvLSTM model [15] is a neural network architecture that combines Convolutional Neural Networks [16] with Long Short-Term Memory [17] networks. It is designed to handle sequences of data that have both spatial and temporal dependencies, making it particularly useful for spatiotemporal data processing and time-series forecasting. 

The convolutional windows of the model are used to recognize shapes and patterns in areas of low vegetation levels post-wildfire. The hidden states and memory of the model are used to learn temporal patterns in vegetation recovery. The ConvLSTM is uniquely designed to handle spatiotemporal information because all inputs $x_1,...,x_t$, cell outputs $c_1,...,c_t$, hidden states $h_1,...,h_t$, and gates $i_t, f_t, o_t$ are 3D tensors where the last two dimensions of each tensor represent its spatial dimensions. Model inputs represent series of images that contain the variables used in model training, while model outputs represent series of images that contain the model's forecasted NDVI values.

In our model architecture, the layer responsible for spatiotemporal learning was the ConvLSTM2D layer. Three such layers were used with filter counts of $32, 128,$ and $64$, and a 3D convolutional layer with a single filter was utilized to produce the final output, a time series of forecasted NDVI ratios. A kernel size of ($3, 3$) was selected for each ConvLSTM2D layer, and a kernel size of ($3, 3, 3$) was selected for the Conv3D layer. The model utilized a zero-padding technique and a convolutional stride size of $(1, 1)$ so that the size of the filtered spatial data was identical to that of the input data. We chose the Rectified Linear Unit (ReLU) function as the activation function for every convolutional layer. Following the first two ConvLSTM2D layers, each tensor was normalized using batch normalization layers, which transform the data along each channel to have a mean output value close to 0 and an output standard deviation close to 1.

\section{Model Training and Results}

\subsection{Training Data}

The input to our ConvLSTM consisted of 5 dimensions, namely samples, timesteps, rows, columns, and channels. Many of the wildfires included in our dataset had short-term re-burns either before or after the post-wildfire analysis period, as well as extreme cases of snow, ice, or cloud cover that disrupted all or most of some timestamps. These wildfires had NDVI ratio values that were extremely erratic and unrepresentative of typical a logistic growth rate, sometimes jumping from average ratio values of 0.5 to 3 or higher in a single timestamp. Such wildfires would be detrimental to model learning, and so we selected $31$ wildfires whose average NDVI ratios showed no erratic increases or decreases to be included in our model's training and validation.

Given the limited number of wildfires, we then partitioned each original 50 x 50 image into 25 sub-images each spanning 10 x 10 pixels. This increased our sample size and reduced the complexity of our learning objective by significantly reducing the size of each sample. Our model was subsequently able to prioritize the accurate prediction of the NDVI ratios for each pixel, rather than identifying the larger shapes of the burned areas.  

With our dataset, we trained our model for 100 epochs, employing 80\% of our filtered data for training and the remaining for validation. We use a learning rate scheduler to change the model's learning rate during training, and selected mean absolute error as our training and validation loss function.

\subsection{Results}

Once the ConvLSTM model was trained, it was tested by running it on randomly selected sample wildfires. The model was provided with 5 consecutive time steps of post-fire observations as input and generated forecasts for the subsequent 20 frames of NDVI data, covering a recovery period of up to 2 years after the fire cessation. Figure \ref{fig:saddle} shows a comparison between the actual and forecasted NDVI values for the Saddle Ridge Fire of 2019. Our ConvLSTM model adeptly captures the overall plant vegetation levels, forecasting the temporal and spatial patterns associated with the data. 

\section{Analysis and Discussion}
\subsection{Plant Growth Curves}

While forecasted NDVI time series provide future vegetation index recovery images, individual NDVI pixels are small and sensitive to random changes and errors in measuring tools. In order to create a general estimator for recovery, tensor regression was chosen to generalize the framework, providing quantified metrics for subsequent analysis and modeling purposes. We chose to use tensor regression to fit the patterns of our post-wildfire recovery to a logistic curve, which is commonly used to describe the growth of plants [18]. Similar to a logistic curve, the rate of vegetation recovery is the fastest immediately after wildfire containment [9]. We first fitted actual post-wildfire NDVI ratios to gain a baseline set of logistic curve parameters for each pixel of each wildfire, namely a growth capacity $L$ and a growth rate $k$. We could then average these values over each wildfire study area and compare those averages with the logistic growth parameters obtained from our fitted model predictions. 

\subsection{Convolutional LSTM Tensor Regression Network}

Upon obtaining these logistic curve parameters, we enhanced our ConvLSTM model by incorporating Tucker tensor regression [19] to efficiently translate NDVI forecasts into measurable logistic growth recovery measurements. This would extend our pixel-wise logistic curve analysis to a higher-dimensional data structure, and our model framework was combined to create a Convolutional LSTM  Tensor Regression (ConvLSTMTR) network. Since tensor regression requires fitting against a collection of tensors along with their target variables' ground truth values, we used the actual $25$ {
\parfillskip=0pt
\parskip=0pt
\par}
\begin{wrapfigure}[18]{L}{0.40\textwidth} 
  \centering
  \includegraphics[width=1.\linewidth]{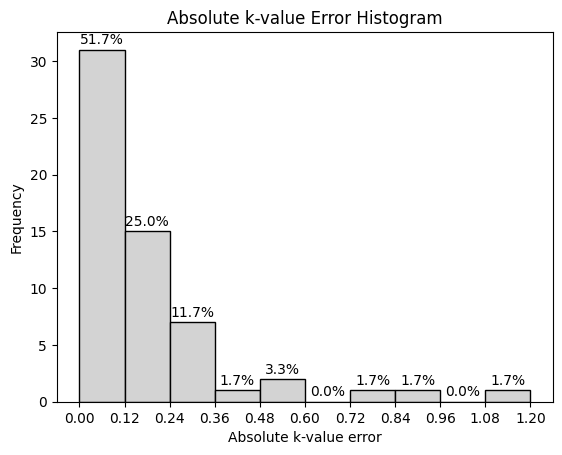}
  \caption{A histogram of out-of-sample absolute differences between k-values of logistic fits on actual NDVI data vs. predicted data.}
  \label{fig:k_error}
\end{wrapfigure}
    frames of NDVI for each subgrid that had a burn area of at least 50\%. We fitted a logistic curve to every pixel in these grids and averaged the fitted values for each grid. With this input, the Tucker tensor regression model could generate a series of estimated logistic curve parameters, providing the final predictions of our ConvLSTMTR.

We analyzed the 73 out-of-sample wildfires and measured the absolute difference between the estimated logistic parameters from tensor regression and those obtained through direct logistic curve fitting on actual NDVI time series data, as shown in Figure \ref{fig:k_error}. Over 50\% of predictions had an absolute error of less than $0.12$, 75\% had an error of less than $0.24$, and 90\% of predictions had an error of less than $0.48$. Outliers occurred in less than 10\% of scenarios. These results showed great promise and demonstrated that our ConvLSTMTR network was not only able to correctly recognize the growth patterns but was also able to accurately adapt to various rates of vegetation recovery.

\subsection{Significance}

Previous research on post-fire plant regeneration has primarily focused on studying plant growth patterns from an ecological and biogeographical perspective, investigating the mechanisms of post-fire plant recovery, and analyzing the impact of various factors on this process. However, there has been a noticeable gap in the literature when it comes to utilizing machine learning models for spatial-temporal forecasting of plant regrowth trajectories. Our research addresses this gap by demonstrating the effectiveness of the ConvLSTM model in accurately predicting future spatial-temporal plant growth patterns, leveraging various global satellite data.

Considering the vulnerability of machine learning models to data quality issues, we integrated KNN imputation to replace bad data. Moreover, we pioneered the application of tensor regression in conjunction with the ConvLSTM, known as the ConvLSTMTR network, to estimate post-wildfire vegetation recovery rates. Our approach combines the strengths of the ConvLSTM in handling high-dimensional spatial-temporal datasets and generating accurate forecasts with the capabilities of tensor regression to extract growth features for interpretation and analysis. As a result, our ConvLSTMTR network exhibits high accuracy in forecasting post-fire recovery rates.

\section{Conclusion}

In this study, we developed a Convolutional Long Short-Term Memory Tensor Regression network (ConvLSTMTR) to predict plant regeneration in areas impacted by wildfires. Approximately 100 wildfires in California from 2013 through 2020 with a burn area of at least 3,000 acres were analyzed, and 26 months of post-fire plant recovery data were used for each wildfire. This data was represented as multi-dimensional spatial-temporal inputs encompassing variables such as NDVI, EVI, a burned area fire mask, precipitation, and surface temperature. To effectively address challenges posed by data quality, limited sample size, and seasonal patterns, we implemented preprocessing techniques including KNN imputation, stratified sampling, and de-seasonalization. Upon completing the training stage, the model was able to forecast NDVI time series 18 months into the future given 5 to 6 months of immediate post-wildfire observations. Following our ConvLSTM forecasting, we used Tucker tensor regression to fit logistic growth and capacity parameters for each predicted NDVI time series. The results from out-of-sample testing indicated that our ConvLSTMTR was able to forecast a logistic growth rate $k$ for each wildfire with impressive performance. Over 50\% of predictions exhibited an absolute error of less than 0.12, and over 75\% had an error of less than 0.24.


\section*{References}

{
\small
[1] National Interagency Fire Center \ (2022) Total wildfires and acres.
https://www.nifc.gov/sites/default/files/document-media/TotalFires.pdf.

[2] Troy, A.\ \& Pusina, T.\ \& Romsos, S.\ \& Modhaddas, J. \& Buchholz, T.\ (2022) The true cost of wildfire in the western U.S. {\it Western Forestry Leadership Coalition},  https://www.thewflc.org/sites/default/files/TrueCostofWildfire.pdf.

[3] California Department of Fish and Wildlife \ (2020) Science: wildfire impacts. https://wildlife.ca.gov/Science-Institute/Wildfire-Impacts.

[4] Shive, K. L.\ \&  Preisler, H.K.\ \& Welch, K.R.\ \& Safford, H.D.\ \& Butz, R.J. \ \& O'Hara, K.L.\ \& Stephens, S.L.\ (2018) From the stand scale to the landscape scale: predicting the spatial patterns of forest regeneration after disturbance. {\it Ecological Applications} {\bf 28}(6):1626-1639. https://doi.org/10.1002/eap.1756.

[5] Davis, K.T.\ \&  Robles, M.D.\ \&  Kemp, K.B.\ \&  Campbell, J.L\ (2023) Reduced fire severity offers near-term buffer to climate-driven declines in conifer resilience across the western United States. {\it Proceedings of the National Academy of Sciences of the United States of America} {\bf 120}(11). https://doi.org/10.1073/pnas.2208120120.

[6] Moya, D.\ \& González-De Vega, S.\ \& Lozano, E.\ \& García-Orenes, F.\ \& Mataix-Solera, J.\ \& Lucas-Borja, M.E.\ \& de las Heras, J.\ (2019) The burn severity and plant recovery relationship affect the biological and chemical soil properties of Pinus halepensis Mill. stands in the short and mid-terms after wildfire. {\it Journal of Environmental Management} {\bf 235}:250-256. https://doi.org/10.1016/j.jenvman.2019.01.029.

[7] Scaven, V.L.\ \& Rafferty, N.E.\ (2013) {Physiological effects of climate warming on flowering plants and insect pollinators and potential consequences for their interactions. {\it Current Zoology} {\bf 59}(3):418-426. https://doi.org/10.1093/czoolo/59.3.418.

[8] Liu, Q.\ \& Fu, B.\ \& Chen, Z.\ \& Chen, L.\ \& Liu, L.\ \& Peng, W.\ \& Liang, Y.\ \& Chen, L.\ (2022) Evaluating effects of post-fire climate and burn severity on the early-term regeneration of forest and shrub communities in the San Gabriel mountains of California from Sentinel-2(MSI) images. {\it Forests} {\bf 13}(7). https://doi.org/10.3390/f13071060.

[9] Bright, B.C.\ \& Hudak, A.T.\ \& Kennedy, R.E.\ \& Braaten, J.D.\ \& Khalyani, A.H.\ (2019) Examining post-fire vegetation recovery with Landsat time series analysis in three western North American forest types. {\it Fire Ecology} {\bf 15}(8). https://doi.org/10.1186/s42408-018-0021-9.

[10] Kondylatos, S.\ \& Prapas, I.\ \& Ronco, M. \ \& Papoutsis, I.\ \& Camps-Valls, G.\ \& Piles, M.\ \& Fernández-Torres, M.\ \& Carvalhais, N.\ (2022) Wildfire danger prediction and understanding with deep learning. {\it AGU Advancing Earth and Space Science} {\bf 49}(17). https://doi.org/10.1029/2022GL099368.

[11] Didan, K.\ (2021) MODIS/Terra vegetation indices 16-day L3 global 250m SIN grid V061. {\it NASA EOSDIS Land Processes Distributed Active Archive Center}. https://doi.org/10.5067/MODIS/MOD13Q1.061.

[12] Collins, C.G.\ \& Elmendorf, S.C.\ \& Hollister, R.D.\ \& Henry, G.H.R\ \& et al.\ (2021) Experimental warming differentially affects vegetative and reproductive phenology of tundra plants. {\it Nature Communications} {\bf 3442}. https://doi.org/10.1038/s41467-021-23841-2.

[13] Wan, Z.\ \& Hook, S.\ \& Hulley, G.\ (2015) MOD11A2 MODIS/Terra land surface temperature/emissivity 8-day L3 global 1km SIN grid V006. {\it NASA EOSDIS Land Processes Distributed Active Archive Center}. https://doi.org/10.5067/MODIS/MOD11A2.006.

[14] Huffman, G.J.\ \& Stocker, E.F.\ \& Bolvin, D.T.\ \& Neklin, E.J.\ \& Tan, J.\ (2019) GPM IMERG early precipitation L3 1 day 0.1 degree x 0.1 degree V06. {\it Goddard Earth Sciences Data and Information Services Center}. https://doi.org/10.5067/GPM/IMERGDE/DAY/06

[15] Shi, X.\ \& Chen, Z\ \& Wang, H.\ \& Yeung, D.-Y.\ \& Wong, W.-K.\ \& Woo, W.-C.\ (2015) Convolutional LSTM network: a machine learning approach for precipitation nowcasting. https://doi.org/10.48550/arXiv.1506.04214

[16] LeCun, Y.\ \& Bottou, L.\ \& Yoshua, B.\ \& Haffner, P.\ (1998) Gradient-based learning applied to document recognition. {\it Proceedings of the IEEE} {\bf 86}(11):2278-2324. https://doi.org/10.1109/5.726791.

[17] Hochreiter, S.\ \& Schmidhuber, J.\ (1997) Long short-term memory. {\it IEEE Xplore Neural Computation} {\bf 9}(8):1735-1780. https://doi.org/10.1162/neco.1997.9.8.1735.

[18] Kawano, T.\ \& Wallbridge, N.\ \& Plummer, C.\ (2020) Logistic models for simulating the growth of plants by defining the maximum plant size as the limit of information flow. {\it Plant Signaling \& Behavior} {\bf 15}(2). https://doi.org/10.1080/15592324.2019.1709718.

[19] Kolda, T.G.\ \& Bader, B.W.\ (2009) Tensor decompositions and applications. {\it Journal of Neuroscience} {\bf 51}(3):455–500. https://doi.org/10.1137/07070111X.


\section{Appendix}

\subsection{K-Nearest Neighbor Clustering} 
Once each wildfire had a set of forecasted logistic growth values, we clustered all wildfires across California to observe any possible patterns. For this, we chose to use K-nearest neighbor (KNN) clustering enhanced with Uniform Manifold Approximation and Projection (UMAP).

\subsubsection{Uniform Manifold Approximation and Projection}

Uniform Manifold Approximation and Projection (UMAP) is a learning technique that can be used to visualize high-dimensional datasets in lower dimensions. Before UMAP, one of the most widely-used techniques for data visualization was t-distributed stochastic neighbor embedding (t-SNE), which constructs high-dimension data in lower dimensions using Euclidean distance to determine similarity between the dimensional maps. UMAP is functionally similar, with a major difference being that UMAP utilizes Riemannian distance rather than Euclidean distance. UMAP first calculates probability distributions over pairs of objects in the high-dimension map, then assigns similar probability distributions to a lower-dimension map of objects. 
\parfillskip=0pt
\parskip=0pt
\par}
\begin{wrapfigure}[19]{R}{0.4\textwidth}
  \centering
  \includegraphics[width=1.\linewidth]{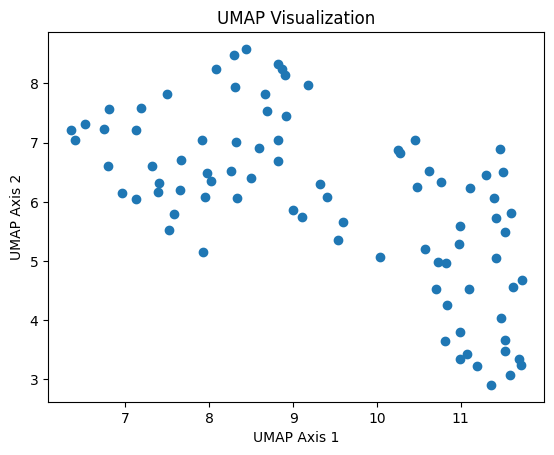}
  \caption{The results of UMAP after projecting our 5D points representing each wildfire into a 2D plane.}
  \label{fig:umap_uncluster}
\end{wrapfigure}
To prepare our data for UMAP, we treated each wildfire as a single point in a 5D plane, where the values of five statistics collected represented their coordinates in the plane: longitude, latitude, average predicted growth rate $k$, average predicted carrying capacity $L$, and average precipitation during the first 5 time steps $p$. Each set of values was also normalized to eliminate variation in feature ranges that could introduce unwanted bias during clustering. Then, we used UMAP to project those points in a 2D plane. Figure \ref{fig:umap_uncluster} shows the projected results generated by UMAP. Note that the x- and y-axes created by UMAP have no statistical significance in relation to the variables represented in the original 5D data.

\subsubsection{Clustering Results}

The obtained UMAP points were then clustered using KNN with $3$, $5$, and $10$ clusters. The raw clustered UMAP results and the cluster labels plotted back on a map of California are displayed in Figure \ref{fig:umap_clusters}. The size of the dots represents the average magnitude of the $k$-values in each cluster. The clustered labels were manually inspected, and two groups of wildfires stood out as always being placed in the same cluster, regardless of the number of clusters being created, indicating that the wildfires in each set were extremely similar to each other in location and in recovery pattern.

\parfillskip=0pt
\parskip=0pt
\par}
\begin{wrapfigure}[19]{L}{0.35\textwidth}
  \centering
  \includegraphics[width=1.\linewidth]{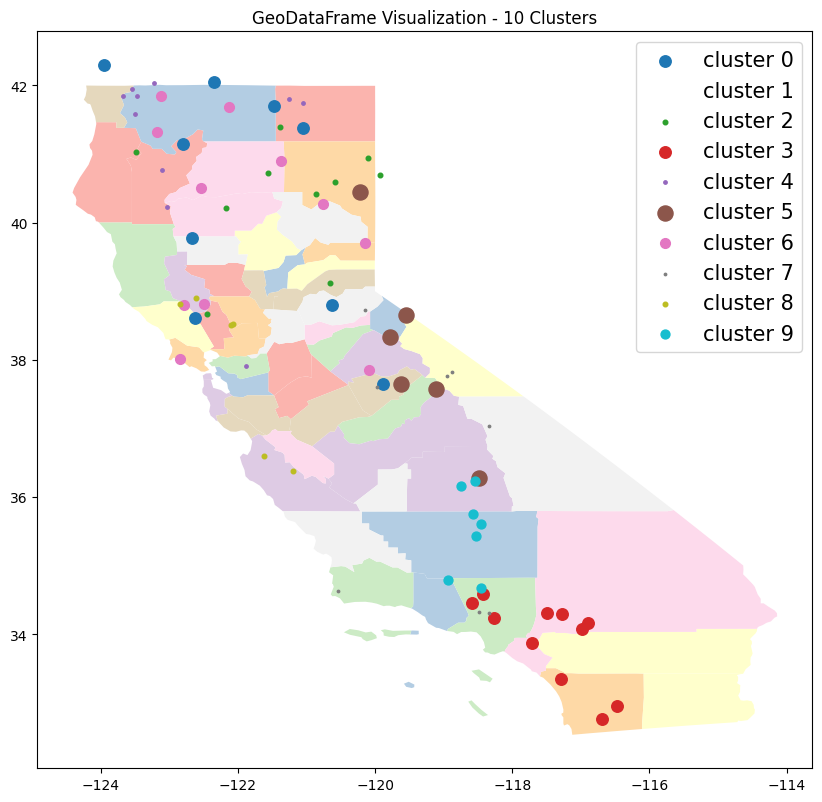}
  \caption{The results of KNN clustering using 10 clusters, plotted on a map of California.}
  \label{fig:umap_clusters}
\end{wrapfigure}

The group of wildfires highlighted in red around the Counties of Los Angeles and San Bernardino were all areas that had large positive $k$-values, signifying that the vegetation in these areas was able to quickly recover back to pre-wildfire levels. This cluster included the Canyon Fire 2 of 2017, which had the highest $k$-value observed in our dataset of $0.611$. On the contrary, the wildfires highlighted in brown around the Counties of Tuolumne and Mariposa were all areas that had large negative $k$-values. This indicates that the vegetation in the area continued to deteriorate , and that these areas could need increased manual intervention to allow vegetation levels to recover. This cluster also contained the Summit Complex Fire of 2017, which had the lowest $k$-value observed in our study of $-1.038$.

\end{document}